\documentclass[10pt,twocolumn]{article}

\usepackage{times}
\usepackage{epsfig}
\usepackage{graphicx}
\usepackage{amsmath}
\usepackage{amssymb}
\usepackage{caption}

\usepackage[pagebackref=true,breaklinks=true,letterpaper=true,colorlinks,bookmarks=false]{hyperref}
\usepackage{amsfonts}
\usepackage{mathrsfs,algorithm,algorithmicx,algpseudocode}
\usepackage{diagbox}
\usepackage{subfigure}
\usepackage{geometry}

\makeatletter
\newcommand{\printfnsymbol}[1]{%
	\textsuperscript{\@fnsymbol{#1}}%
}
\makeatother

\begin{document}

\title{GDRQ: Group-based Distribution Reshaping for Quantization}

\author{
	Haibao Yu$^1$\thanks{equal contribution}~~~
	Tuopu Wen$^2$\printfnsymbol{1}~~~
	Guangliang Cheng$^1$~~~
	Jiankai Sun$^3$~~~
	Qi Han$^1$~~~
	Jianping Shi$^1$~~~
	\\
	\and
	$^1$SenseTime Research~~~~~$^2$Tsinghua University~~~~ $^3$The Chinese University of Hong Kong\\$^4$Beihang Univerisity\\
	\and
	{\tt\small 
		\{yuhaibao, chengguangliang, shijianping\}@sensetime.com~~~wtp18@mails.tsinghua.edu.cn}\\
   {\tt\small 
       sj019@ie.cuhk.edu.hk~~~ dismilk@buaa.edu.cn}
}
\date{}
\maketitle

	\begin{abstract}
		Low-bit quantization is challenging to maintain high performance with limited model capacity (e.g., 4-bit for both weights and activations).  
		Naturally, the distribution of both weights and activations in deep neural network are Gaussian-like.
		Nevertheless, due to the limited bitwidth of low-bit model, uniform-like distributed weights and activations have been proved to be more friendly to quantization while preserving accuracy~\cite{Han2015Learning}. 
		Motivated by this, we propose Scale-Clip, a Distribution Reshaping technique that can reshape weights or activations into a uniform-like distribution in a dynamic manner.
		Furthermore, to increase the model capability for a low-bit model, a novel Group-based Quantization algorithm is proposed to split the filters into several groups. Different groups can learn different quantization parameters, which can be elegantly merged in to batch normalization layer without extra computational cost in the inference stage. 
		Finally, we integrate Scale-Clip technique with Group-based Quantization algorithm and propose the Group-based Distribution Reshaping Quantization (GDQR) framework to further improve the quantization performance.
		Experiments on various networks (e.g. VGGNet and ResNet) and vision tasks (e.g. classification, detection and segmentation) demonstrate that our framework achieves much better performance than state-of-the-art quantization methods. 
		Specifically, the ResNet-$50$ model with $2$-bit weights and 4-bit activations obtained by our framework achieves less than $1\%$ accuracy drop on ImageNet classification task, which is a new state-of-the-art to our best knowledge.
	\end{abstract}

\section{Introduction}
\begin{figure}
	\centering
	\includegraphics[width=0.45\textwidth]{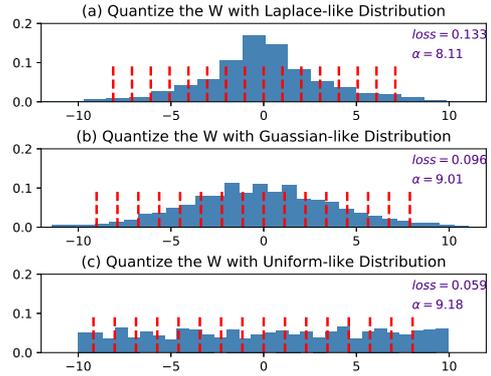}
	\caption{We respectively quantize the $\mathbf{W}$s which obey: (a) Laplace-like distribution, (b) Gaussian-like distribution and (c) Uniform-like distribution into $Q(\mathbf{W})$ as Eq. \ref{eq: min Q} with $n_w=4$ and $||\cdot||_p=||\cdot||_1$, where red lines means the quantization bins. We calculate \textit{quantized-loss} as Eq. \ref{quantized-loss-redefine}. Comparatively, uniform distribution better fits the uniform quantization.}\label{fig:different_distribution}
\end{figure}
\par In recent years, convolutional neural networks (CNNs) have achieved significant breakthroughs in a variety of computer vision tasks, 
such as image classification~\cite{deng2009imagenet,he2016deep}, object detection~\cite{ren2015faster,girshick2015fast,redmon2016you}, and semantic segmentation~\cite{zhao2017pyramid,long2015fully}, etc.
These deep neural networks are usually computational-intensive and resource-consuming, which restricts them to be deployment on resource-limited devices (\emph{e.g.}, ARM and FPGA).
In order to improve the hardware efficiency, many researchers have proposed to quantize the weights and activations into low-bit~\cite{han2015deep,Zhu2016Trained}, especially in a linear way such as uniform quantization.

Nevertheless, quantization results in performance degradation inevitably because of the indifferentiability and limited expression capacity of the deep neural networks. Previous methods adopt different strategies to alleviate the performance degradation.
Given that the indifferentiability of the quantized feature expression, 
much effort have been invested into developing efficient quantization training frameworks~\cite{zhou2016dorefa,gysel2016ristretto,zhou2017incremental,denton2014exploiting,mishra2017apprentice,wang2018two,courbariaux2016binarized}, 
such as conducting forward propagation in low-bit mode while backward propagation with latent full precision weights, 
or minimizing the KL-divergence between the original weights and quantized weights when training. 
Such methods do not perform well for low-bit (e.g. $2$-bit) situations due to the limited expression capacity.
To further enhance low-bit expression capacity, non-uniform quantization is proposed ~\cite{park2017weighted,zhou2017balanced,zhou2017incremental,miyashita2016convolutional}, 
such as remapping each bit via a logarithm manner or maximizing the quantized weight entropy. 
Following the natural Gaussian distribution of weights and activations, 
this method requires more complex hardware deployment process.
%
To better utilize the low-bit expression capacity,  we propose a new Distribution Reshaping method named Scale-Clip for uniform quantization. 
We analyze the natural distribution of weights and activations,
and found most of them are Gaussian-like or Laplace-like.
Compared to uniform distribution, these distributions cause more \emph{quantized-loss} and are not suitable for uniform quantization, which are shown  in
Fig. \ref{fig:different_distribution}.
It is rational to initialize the model with uniform distribution  for uniform quantization, which has been discussed in ~\cite{han2015deep}.
Further, it is necessary to provide a pre-training model for uniform quantization. This pre-training model not only has high precision, but also has a uniform distribution, so that the quantitative model can be stably restored to a high-precision state.
Therefore, our proposed Scale-Clip method applies the Distribution Reshaping constraint to fit the distribution of weights and activations to quantized uniform while not affecting the training of the pre-training model.
Another challenge is to enhance the capacity of low-bit model. 
A widely-adopted approach is to quantize the weights and activations linearly and let the weights and activations of each layer share the same linear quantization factor.
However, we observe that filters in the $k^{th}$ layer are not necessarily to share the same quantization bins.
It is rational to use flexible quantization bins for different filters.
Thus we propose the Group-based Quantization (GQ) that clusters the filters into groups for quantizing.
GQ allows a more flexible way for quantized weights to take quantization bins, resulting in enhancing the low-bit model's capacity, 
Distribution Reshaping method can also be applied into group filters. 
Incorporating the Distribution Reshaping method into Group-based Quantization, 
we propose Group-based Distribution Reshaping Quantization (GDRQ) framework 
that reshapes the weights and activations of each group filters into more uniform-like distribution for quantizing.
	Our GDRQ framework has the following advantages.
	(1) Models directly use uniform quantization expression, which is easy to be deployed on resource-limited devices.
	(2) Our proposed Distribution Reshaping method can optimize the original distribution of weights and activations more quantized uniform, which fully utilizes the capacity of low-bit representation while retains performance. 
	(3) Our proposed Group-based Quantization can enhance the low-bit model's capacity while not impact the deployment.
	(4) Our framework is generally useful to all vision tasks with different network complexity.\
	The main contributions of this paper can be summarized as follows:
	
	\begin{enumerate}
		\item \textbf{Distribution Reshaping:} 
		We propose a simple yet effective Distribution Reshaping method based constraint to shape the distribution of floating-point model to be uniform-like, which is more suitable for quantization.
		\item \textbf{Group-based Quantization:} We extend the quantization into Group-based quantization to increase the low-bit model's capacity.
		\item \textbf{GDRQ framework:} We incorporate the Distribution Reshaping method and Group-based quantization into our quantization framework, and validate that our framework outperforms state-of-art methods in a variety of networks and tasks. 
	\end{enumerate}

\begin{figure*}
	\centering
    \begin{minipage}[b]{0.9\linewidth} 
    \centering   
    \includegraphics[width=0.8\linewidth]{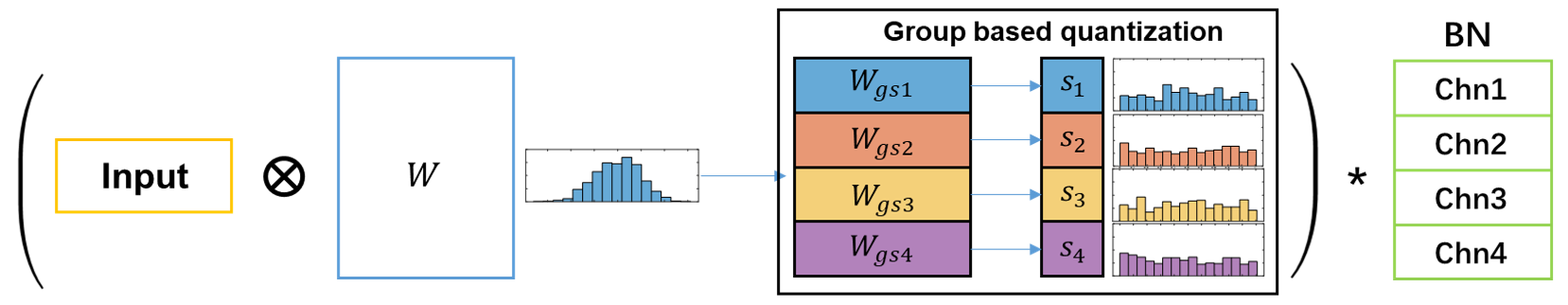}
    \caption*{(a) Group-based Distribution Reshaping Quantization}
    \end{minipage}
    \begin{minipage}[b]{0.9\linewidth} 
    \centering
    \includegraphics[width=0.8\linewidth]{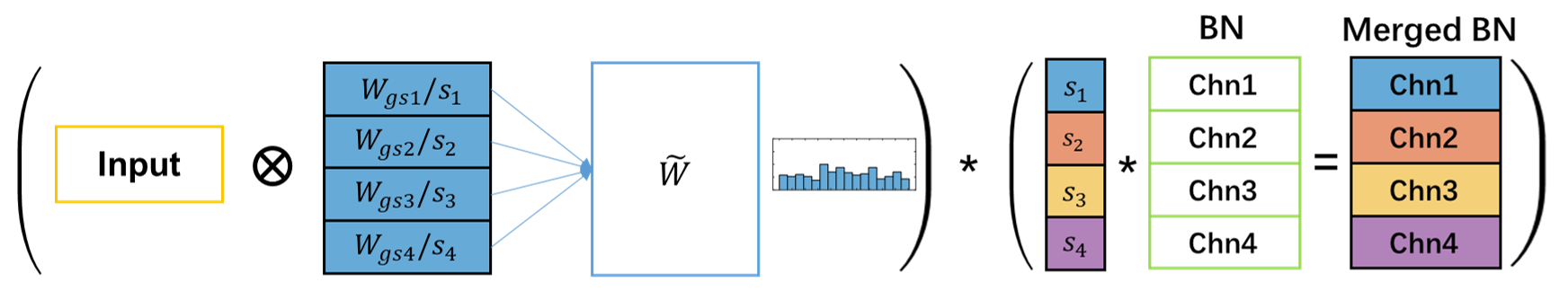}
    \caption*{(b) Inference for Group-based Distribution Reshaping Quantization}
    \end{minipage}   
	\caption{Overview of our quantization framework. (a) illustrates the main flow of proposed group-based quantization. Weights are divided into several groups and their distributions are respectively clipped with different thresholds $T^w_i$ and reshaped into uniform-like distributions. Then the uniform quantization is performed on the reshaped distribution of each weight group.
	In (b), during test phase, different clipping thresholds for each group can be merged into following batch normalized layer.}
	\label{fig:overview}
\end{figure*}

	\section{Related Work}
Convolution neural networks have achieved remarkable performance and have been widely used in a variety of computer vision tasks.
To deploy the CNN models on resource-limited devices (\emph{e.g.}, mobile phones or self-driving cars),
many model compression algorithms~\cite{cheng2018recent,Sze2017Efficient} have been proposed to reduce the model's storage as well as to accelerate inference.

	\paragraph{Quantization}
	Quantization can be used for reducing the number of bits required to represent weights and activations. Quantization techniques can be roughly categorized into non-uniform quantization and uniform quantization. 	
	Non-uniform quantization usually contains scalar and vector quantization. \cite{miyashita2016convolutional,zhou2017incremental} quantize the network as logarithmic numbers. 
	BalanceQ~\cite{zhou2017balanced} selects the quantization bins by Histogram Equalization while~\cite{park2017weighted} takes weighted entropy as the measurement.~\cite{wu2016quantized} regards convolution and full-connected layers as inter product operations and thus transfer the product quantization into the network quantization. 
	FFN~\cite{wang2017fixed} approximates weight matrices using the weighted sum of outer product of several vector pairs with ternary entries vectors, which facilitates the network deployment on fixed-point computation architectures.
	Most of the above methods requires more bits to represent numbers during arithmetic computation, making it inconvenient to be deployed on resource-limited devices.	
	Uniform quantization is more hardware friendly. Researches are focusing on designing effective quantization training framework to deal with the indifferentiability of quantization. 
	Previous works (\emph{e.g.},DoReFa-Net~\cite{zhou2016dorefa}) utilize straight-through estimator (STE) to estimate the quantization gradient. Ristretto~\cite{gysel2016ristretto} proposes to calculate gradients with quantized parameters while updating the gradients on the latent floating-point weights. 
	Some works~\cite{courbariaux2016binarized,rastegari2016xnor,mellempudi2017ternary,wang2018two} focus on extremely low-bit quantization training strategy and obtain quantization levels by minimizing reconstruction error. 
	ELQ~\cite{zhou2018explicit} adopts incremental training strategy, which fixes part of weights and update the rest to compensate the degradation of performance. 
	HWGQ~\cite{cai2017deep} introduces clipped and log-tailed ReLU versions to remove outliers and utilizes Half Wave Gaussian Quantizer to optimize the quantization intervals of activations. 
	Based on clipped ReLU, PACT~\cite{choi2018pact} further adaptively learns and determines the clipping parameter $\alpha$ during training for uniform quantization. 
	There are other recent work~\cite{yin2019understanding, hou2019analysis} that theoretically reveals the advantages of clipped ReLU in training quantitative models.
	Currently, there are also some interesting works like HAQ and MPQ~\cite{wu2018mixed,wang2018haq} focusing on how to search the proper bit for the weights and activations with the help of reinforcement learning.
	Although great progress has been made in uniform quantization approaches, the non-negligible performance decrease in large scale dataset still exists.

\section{Method}

To compress the model and deploy it on resource-limited devices, 
we propose a Group-based Distribution Reshaping Quantization (GDRQ) framework for low-bit uniform quantization.
This framework consists of two strategies:
(1) Distribution Reshaping ($\mathbf{DR}$) for quantization,
(2) Group-based Quantization ($\mathbf{GQ}$).
The two methods solve two major challenges for low-bit uniform quantization respectively:
(1) Hard to train due to the discontinuity, 
(2) Weak expression capacity.
The overview of the proposed framework is shown as Fig. \ref{fig:overview}.

\subsection{Preliminary}
Before presenting the detailed framework, some preliminary knowledge of uniform quantization are introduced. We denote the convolutional weights as $\mathbf{W}=\{\mathbf{W}_i|i=1,\cdots,n\}$. 
For each weight atom $w\in \mathbf{W}_i$, 
uniform quantization linearly discretizes it as Eq. \ref{uniform-quantization}.
\begin{equation}\label{uniform-quantization}
	Q(w;\alpha)=[\frac{\text{clamp}(w,\alpha)}{s}]\cdot s,
\end{equation}
where $\text{clamp}(\cdot,\alpha)$ is to truncate the values into $[-\alpha, \alpha]$, $[\cdot]$ is the rounding operation and $\alpha$ is the clipping value. 
The scaling factor $s$ is defined as Eq. \ref{equ:scaling-factor}.
\begin{equation}\label{equ:scaling-factor}
    s(\alpha)=\frac{\alpha}{2^{n_w-1}-1}
\end{equation}
For activations, the uniform quantization truncates the values into the range $[0,\alpha]$ since the activation values are non-negative after the ReLU layer. 
For brevity, we respectively denote the weight quantization and activation quantization as  $Q(\mathbf{W};\alpha)$ and $Q(\mathbf{A};\alpha)$.

\subsection{Distribution Reshaping (DR)}
\paragraph{Motivation}
Low-bit models often confront the problems such as unstable gradient or accuracy dropping during training stage, which is partly caused by the discontinuity of $Q(\cdot;\alpha)$. To solve this problem, some existing works focus on searching an appropriate $Q(\cdot;\alpha)$:
inspired by the formulation of signal-to-noise ratio (\emph{SNR}), widely adopted works directly optimize the \emph{quantized-loss}, 
which is defined as Eq. \ref{quantized-loss-redefine}.
\begin{equation}\label{quantized-loss-redefine}
	QL(\mathbf{W},Q(\mathbf{W};\alpha))=\frac{||\mathbf{W}-Q(\mathbf{W};\alpha)||_p}{||\mathbf{W}||_p}.
\end{equation}
where $||\cdot||_p$ denotes the p-norm and here we take $||\cdot||_p$ as $||\cdot||_1$. 
We can obtain the optimal $\alpha$ as Eq. \ref{eq: min_Q_1}.
\begin{equation}\label{eq: min_Q_1}
    \alpha^{*}=\min_{\alpha}\frac{||\mathbf{W}-Q(\mathbf{W};\alpha)||_p}{||\mathbf{W}||_p}
\end{equation}

Based on the above works, we focus on optimizing the whole $QL(\mathbf{W},Q(\mathbf{W};\alpha))$.
However, due to the indifferentiability of $Q(\cdot;\alpha)$, Eq.~\ref{quantized-loss-redefine} can not be directly optimized during training process.
We resort to search what kind of $\mathbf{W}$ fits the uniform quantization while maintaining the performance.
We observe that most of the weights in convolutional layers distribute near zero areas, (\emph{e.g.}, Laplace distribution or Gaussian distribution). 
However, these distributions often produce large \emph{quantized-loss} compared to uniform distribution.
In Fig.~\ref{fig:different_distribution},
we respectively generate three data distribution examples ((a) Laplace distribution, (b) Gaussian distribution and (c) uniform distribution) composed of 1000 samples. The optimal $\alpha^{*}$ for (a), (b) and (c) are calculated as $\alpha^{*}_a=8.11$, $\alpha^{*}_b=9.01$ and $\alpha^{*}_c=9.18$ according to Eq.~\ref{eq: min_Q_1}. The corresponding \emph{quantized-loss} of uniform distribution is 0.059 while \emph{quantized-loss} about Laplace distribution reaches 0.133.
We can see that the flatter the $\mathbf{W}$ is,  the smaller the \emph{quantized-loss} is.

\begin{figure}
	\centering
	\includegraphics[width=0.45\textwidth]{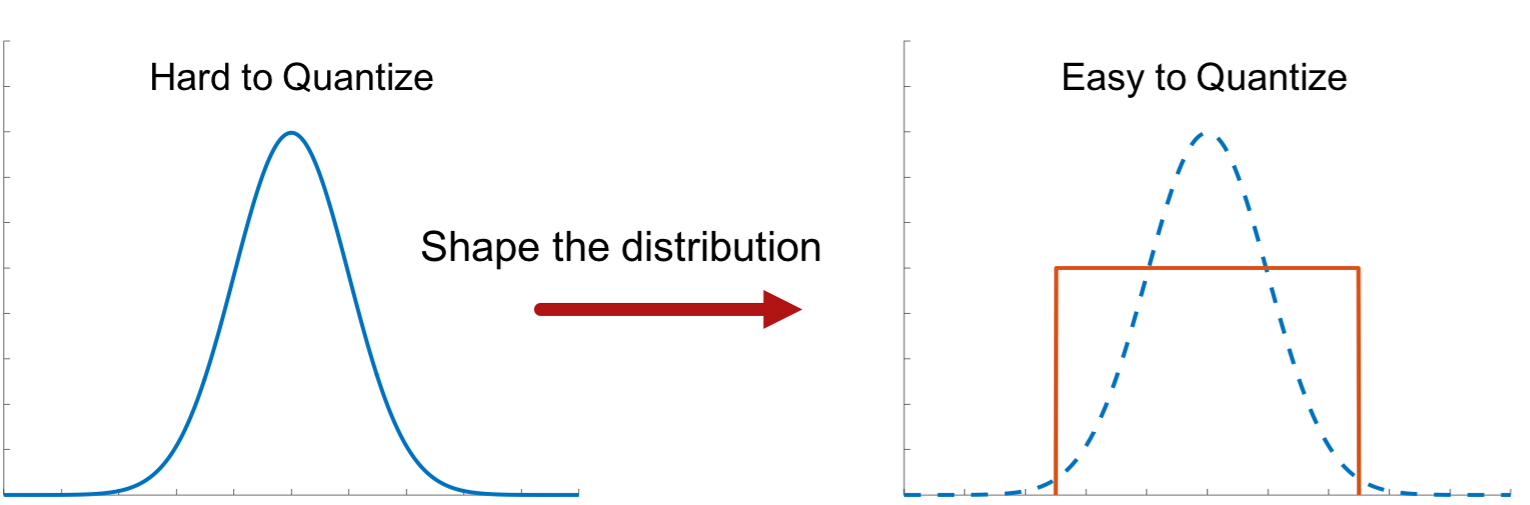}
	\caption{Shape the Gaussian-like distribution into uniform-like distribution for low-bit uniform quantization}\label{fig:reshape_uniform}
\end{figure}

To further verify this conclusion, we have experimentally proved that reshaping convolutional weights with uniform distribution doesn't affect the performance of floating-point model, please refer to the result in Section~\ref{Exp: distribution reshaping}. Thus, we reshape the distribution of floating-point model into uniform distribution, as Fig.~\ref{fig:reshape_uniform} illustrated.

Note that we can also rewrite Eq. \ref{quantized-loss-redefine} in KL divergence and achieve similar results.
And similar works could also seen in ~\cite{wang2018haq, faraone2018syq}.
The difference is that these works are based on the full-precision model to search for the quantization factor to optimize the SNR or KL divergence, and our optimization is divided into two steps, that is, optimizing the full-precision model firstly, then searching for the quantization factor.

\paragraph{Implementation}
$\max(|\mathbf{W}|)$ and $\text{mean}(|\mathbf{W}|)$ are two widely used statistical measures of $\mathbf{W}$.
To start with, we explore the relationship between the two statistical measures under the uniform distribution,
whose density function is defined as Eq. \ref{equ:density-function}.
\begin{equation}\label{equ:density-function}
p(w)=\begin{cases}
C,\quad w\in[-T, T] \\
0,\quad else
\end{cases} 
\end{equation}
where $C=\frac{1}{2T}$.
Suppose $\mathbf{W}$ follows uniform distribution in $[-T,T]$, $\max(|\mathbf{W}|)=T$.
Then $\text{mean}(|\mathbf{W}|)$ can be approximated as Eq. \ref{equ:max-mean-appr}.
\begin{equation}\label{equ:max-mean-appr}
\begin{aligned}
	\text{mean}(|\mathbf{W}|)& \approx \int p(w)|w|dw \\
	                              & = \int_{-T}^{T}\frac{1}{2T}|w|dw = \frac{T}{2}\\
\end{aligned}
\end{equation}
Thus we have T as Eq. \ref{equ:T}.
\begin{equation}\label{equ:T}
	T = \max(|\mathbf{W}|) \approx 2 \cdot \text{mean}(|\mathbf{W}|)
\end{equation}

Based on this relationship, we provide a simple yet effective layer Distribution Reshaping method, 
to reshape the distribution of a floating-point model into uniform distribution dynamically during training stage, 
which has the formulation as Eq. \ref{scale-clip}:
\begin{equation}
\label{scale-clip}
\text{clip}(w)=\left\{
\begin{array}{lr}
T^w, & w \ge T^w \\
w, & w \in (-T^w, T^w) \\
-T^w, & w \le -T^w
\end{array}
\right.
\end{equation}
where
\begin{equation}\label{scale-clip-threshold}
T^w=k \cdot \text{mean}(|\mathbf{W}|).
\end{equation}
The clipping benefits from the proposed Distribution Reshaping method with the following intuitive analysis: when $k$ is near 2, to compensate the lost energy from clipping outliers, more values around zero tend to become larger values. Eventually the $\mathbf{W}$ reaches the limiting case, that is the distribution of $\mathbf{W}$ tend to be uniform.
However, when $k \ll 2$, more outliers will be clipped while there are not enough shifted values to compensate the lost energy,
resulting in the $\mathbf{W}$ converging to zero.
When $k \gg 2$, the distribution gradually become Gaussian-like and eventually the proposed method will have little impact on distribution reshaping.

Activation $\mathbf{A}$ can also adopt Distribution Reshaping strategy.
Nevertheless, the statistical measures of $\mathbf{A}$ is dependent on the data and unstable in training process. Thus, we can not directly employ Eq.~\ref{scale-clip-threshold} on activation quantization. To handle this, a large $k$ should be chosen to adapt to the changeable statistical measures $\text{mean}(\mathbf{A})$. In addition, to achieve stable quantization, we introduce a new update strategy of $T^a$ in training process as Eq. \ref{scale-clip-activation_2} to dynamically satisfy Eq. \ref{scale-clip-activation-threshold}.
\begin{equation}
\label{scale-clip-activation_2}
\begin{aligned}
	T^a &=T^a+\lambda \triangledown T^a \\
	&=T^a+\lambda (T^a-k \cdot \text{mean}(|\mathbf{A}|)).
\end{aligned}
\end{equation}
\begin{equation}\label{scale-clip-activation-threshold}
T^a=\mathop{\arg\min}_{T} \frac{1}{2}||T-k \cdot \text{mean}(|\mathbf{A}|)||_2^2.
\end{equation}
Therefore, the distribution reshaping method can reshape the distribution of activation as uniform-like distribution while maintaining the performance.

Note that clipped method has already been widely used in training the deep neural network, such as gradient clipping~\cite{abadi2016deep, neelakantan2015adding} for avoiding exploding gradients and activation clipping for training quantization model ~\cite{han2015deep, choi2018pact}.
In our work, we just took advantage of the clipped method as part of our optimization, and compared with activation clipping~\cite{han2015deep, choi2018pact}, we further analyze the reasons for the advantages of the chipped method for uniform quantization and theoretically analyze how to set a reasonable threshold.

\subsection{Group-based Quantization}
In this part, to increase low-bit model's capacity, we propose a novel group-based quantization algorithm that splits the filters $\mathbf{W}$ into several groups, then quantizing the grouped filters by adopting different $\alpha$. 
The group-based quantization is shown in Fig.~\ref{fig:overview}.

\begin{figure}
    \centering
    \includegraphics[width=0.45\textwidth]{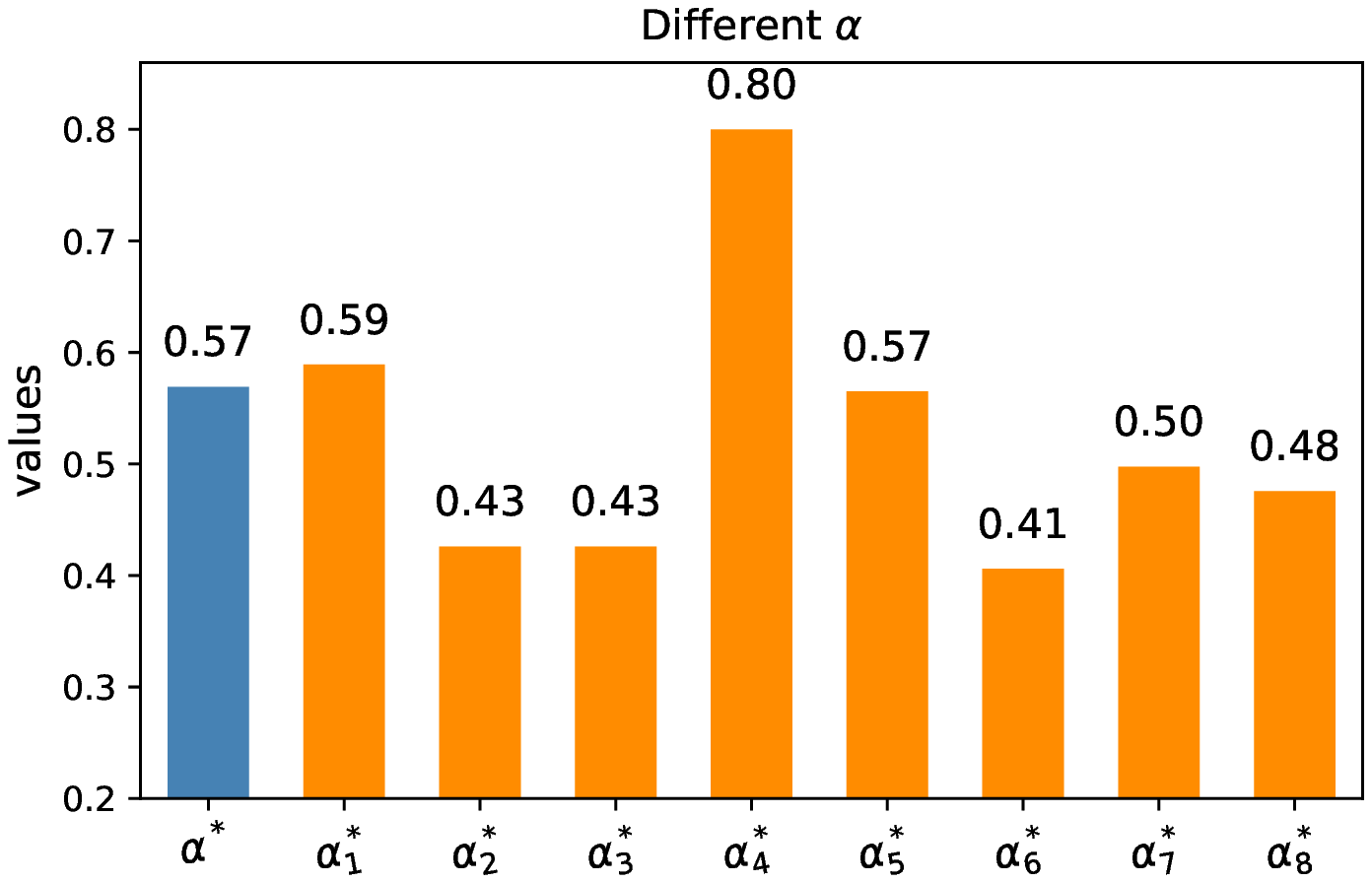}
    \caption{Optimal $\alpha$ for first convolutional layer's weights $\mathbf{W}$ of trained ResNet-$18$. Blue bar represents the optimal $\alpha^{*}$ for $\mathbf{W}$ while orange bars are correspond to each group filters.}
    \label{fig:optimal-alphas}
\end{figure}

The inspiration of the group-based quantization is derived from the observation that quantized weights can only take limited values from $\{\frac{\alpha *i}{2^{(n_w-1)}}| \quad i=-2^{(n_w-1)},\cdots, 2^{(n_w-1)}-1\}$.
 However due to the limited expression capacity, there are some conflicts between the optimal $\alpha$ for all the filters and the optimal $\alpha$s for each group filters. Actually, to achieve better performance, the intuitive solution is that different filters should adopt different $\alpha$ and scaling factor $s(\alpha)$ in the quantization process. For instance, we split the trained weight filters from the first convolutional layer in ResNet-$18$ on CIFAR-$100$ into $8$ groups, and calculate the optimal $\alpha$ for each group filters, respectively. Fig.~\ref{fig:optimal-alphas} illustrates that the optimal $\alpha$ (the blue bar) for all the filters is not always consistent with the optimal $\alpha_l$ (the orange bar) for each group filters. In addition, the $\alpha_l$s for group filters provide strong diversity for quantization, which will enhance the network capacity without extra bit width. As Fig.~\ref{fig:overview} illustrates, the scaling factor $s(\alpha_l)$ for each group filters can be gracefully merged into BN layer. Compared to convolution layer, linear operations in BN layer cost negligible resource in resource-limited devices.
\par Furthermore, quantization with different groups can facilitate reducing the \emph{quantized-loss}.
If we reformulate Eq. \ref{eq: min Q} with 
$QL(\mathbf{W},Q(\mathbf{W};\alpha))=\frac{\sum ||\mathbf{W}_i-Q(\mathbf{W}_i;\alpha)||_1}{||\mathbf{W}||_1},$
and then quantize $\mathbf{W}_i$ with different $\alpha_i$,
we can get that
$QL(\mathbf{W},Q(\mathbf{W};\alpha_1^{*},\cdots,\alpha_n^{*}))\leq QL(\mathbf{W},Q(\mathbf{W};\alpha^{*})),$
where 
$\alpha_i^{*}=\min_{\alpha}||\mathbf{W_i}-Q(\mathbf{W_i};\alpha)||_1/||\mathbf{W}||_1.$
Thus quantizing with different scaling factor $s(\alpha_i)$ promote the low-bit model to be more flexible.

The implementation details of \textbf{Group-based Quantization} are: 
\begin{itemize}
    \item Firstly, decomposing convolution filters $\mathbf{W}$ into group
$\mathbf{G}_l=\{\mathbf{W}_{(l-1)*gs+1},\cdots,\mathbf{W}_{l*gs}\},l=1, \cdots, \frac{n}{gs}$,  
where $gs$ is group size.
    \item Secondly, quantizing the group filters $\mathbf{G}_l$ with  $\alpha_l$ calculated by following form
\begin{equation}\label{eq: min Q}
    \alpha_l^{*}=\mathop{\arg\min}_{\alpha}\frac{||\mathbf{G}_l-Q(\mathbf{G}_l;\alpha)||_1}{||\mathbf{W}||_1},
\end{equation}

\end{itemize}

\subsection{Group-based Distribution Reshaping Quantization Framework}
 The two proposed strategies, distribution reshaping for quantization and group-based quantization, can be integrated together into the quantization framework. In this section, we will introduce the implementation details of Group-based Distribution Reshaping Quantization (GDRQ) framework.
By applying the Distribution Reshaping on each group filters,
we clip the $\mathbf{G}_l$ with
$\alpha_l=k \cdot mean(|\mathbf{G}_l|)$
and reshape the distribution of each group filters as uniform-like.
The training and inference operation are also demonstrated in Fig.~\ref{fig:overview}. Subsequently, the reshaped group filters are quantized with scaling factor $s(\alpha_l)$.
In the inference stage, we merge the scaling factors $s(\alpha_l)$ into BN layers, so that the weights in different groups will share a same uniform quantization range, which is equivalence to traditional quantization setting. 
Therefore we can also easily deploy the quantized low-bit model into resource-limited devices under our GDRQ framework.
The key operations in our quantization framework are illustrated in Alg. \ref{framework-algorithm}.
\begin{algorithm}[ht]
	\caption{Group-based Distribution Reshaping Quantization Framework}
	\label{framework-algorithm}
\begin{algorithmic}
	\Require bit width $n_w$ and $n_a$
	\Ensure Low-bit inference model
	\State Cluster the filters into groups $\mathbf{G}_l$
	\While{Training}
	\For{ each layer}
	\For {each group filters $\mathbf{G}_l$}
	\State Reshape the  $\mathbf{G}_l$ into uniform-like with $T^w_l$
	\State Quantize the group filters into $n_w$-bit
	\EndFor
	\State Reshape the activations into uniform-like with $T^a$
	\State Quantize activations into $n_a$-bit
	\EndFor
	\EndWhile
	\For{each layer and group filters}
	\State Merge the $\alpha_l$ (that is $T^w_l$) into BN layer
	\EndFor
\end{algorithmic}
\end{algorithm}

	\section{Experiment}
	We conduct experiments to validate our proposed Distribution reshaping method and Group-based Quantization in Section \ref{Exp: distribution reshaping} and Section \ref{Exp: group filter}.
	Extensive experiments on varieties of networks and tasks to demonstrate the effectiveness of our GDRQ framework are shown in Section \ref{Exp: Networks}, Section \ref{Exp: detection} and Section \ref{exp:seg}.
	More extensive experiments could be seen in supplementary materials.
	
	\subsection{Distribution Reshaping method validation}\label{Exp: distribution reshaping}
    In this part, we conduct experiments in two steps:
	(1) validating that Distribution Reshaping method can reshape the distribution of weights into different shape, especially when $k_w$ is near $2$ the shape is uniform-like,
	(2) validating that uniform distribution facilitates reducing the \emph{quantized-error} as well as promotes the low-bit model's performance.
    \vspace{-0.5cm}
	\paragraph{Reshaping Effect}
	The experiments are performed on CIFAR-$100$ dataset.
	As our focus is on the validation of Distribution Reshaping method,
	we set different $k_w\in\{2, 2.5, 3, 4, \infty\}$ shown in Eq. \ref{scale-clip-threshold} 
	and impose the Distribution Reshaping method on the convolutional weights 
	to train five floating-point ResNet-$18$ where $k_w=$ means there is no reshaping.
    	
	\begin{figure}[ht!]
		\includegraphics[width=0.46\textwidth]{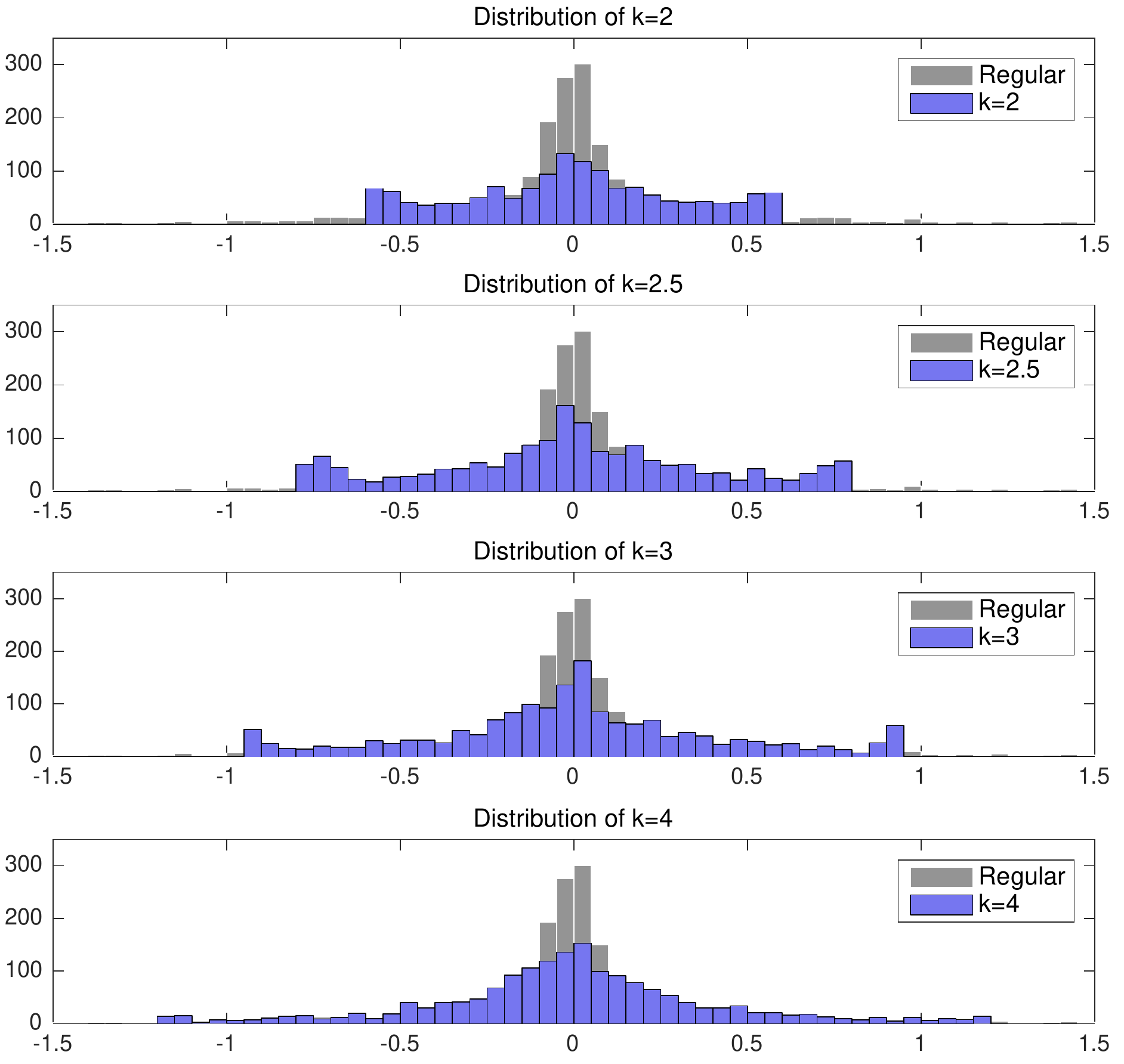}
		\caption{The blue bars show the weight distribution of the first convolutional layer with Distribution Reshaping using different Scale-Clip factors. 
		The gray bars show the weight distribution of the first convolution layer trained without Distribution Reshaping.}\label{CIFAR-$100$_distribution}
	\end{figure}
	
    In Fig. \ref{CIFAR-$100$_distribution}, we present the first convolutional weights' distribution of the five floating-point models.
    As $k_w$ decreases, the distribution becomes flatter with little outliers,
    especially when $k_w = 2$, the distribution becomes almost uniform.
    This phenomenon corresponds to the effectiveness of our Distribution Reshaping method.
    \vspace{-0.3cm}
    \paragraph{Performance Comparison}
    We quantize all convolutional layers' weights of above five floating-point ResNet18 into $n_w$-bit (from $2$-bit to 8-bit),
    and compute the \emph{quantized-loss} of first convolutional layers' weights as Eq. \ref{quantized-loss-redefine}. 
    	
\begin{figure}	
  \centering
    \vspace{-0.3cm}
  \subfigure[Quantized-loss]{
    \label{fig:qwlr} 
    \includegraphics[width=0.48\linewidth]{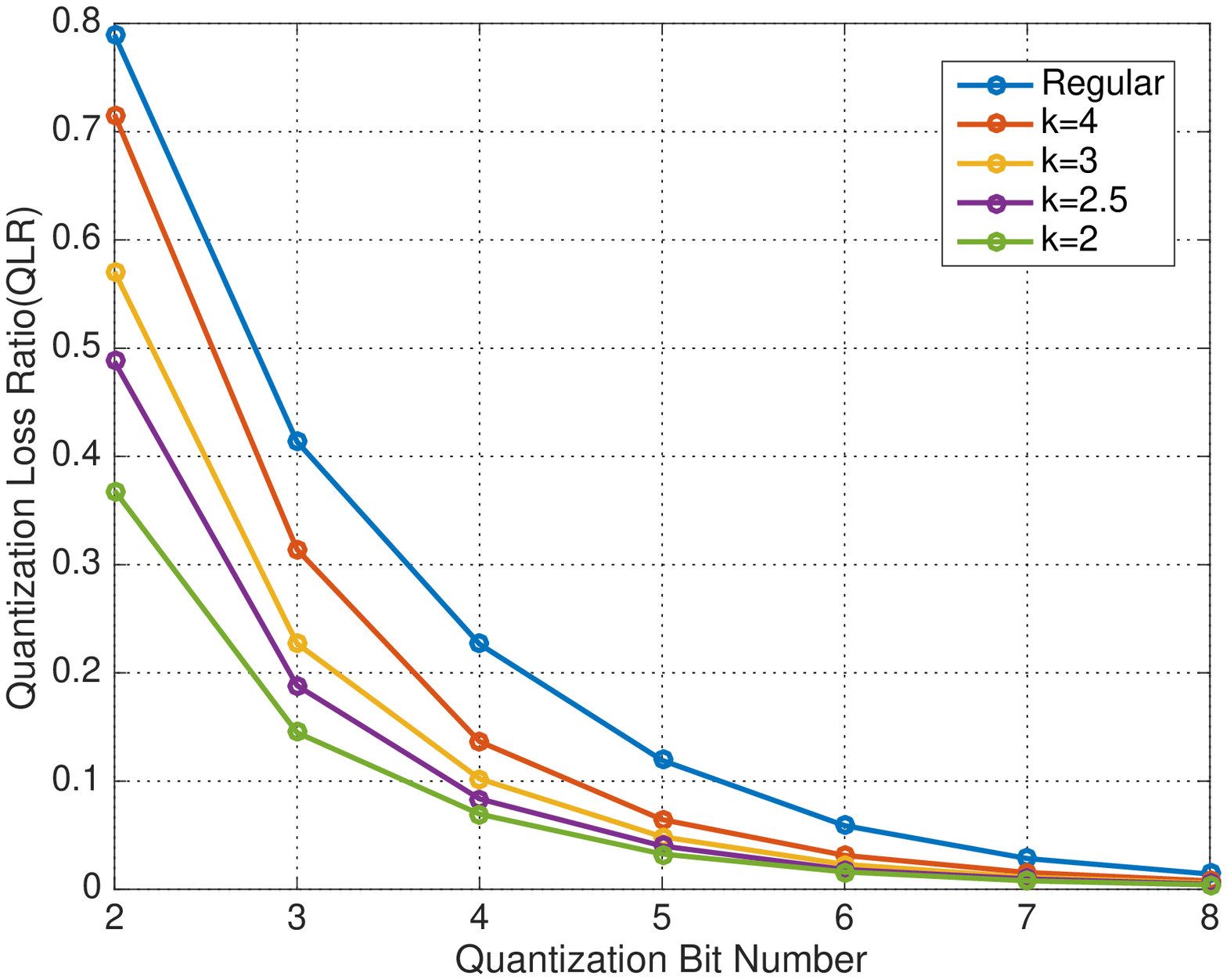}}
  \subfigure[Top-$1$ acurracy]{
    \label{fig:diffk} 
    \includegraphics[width=0.45\linewidth]{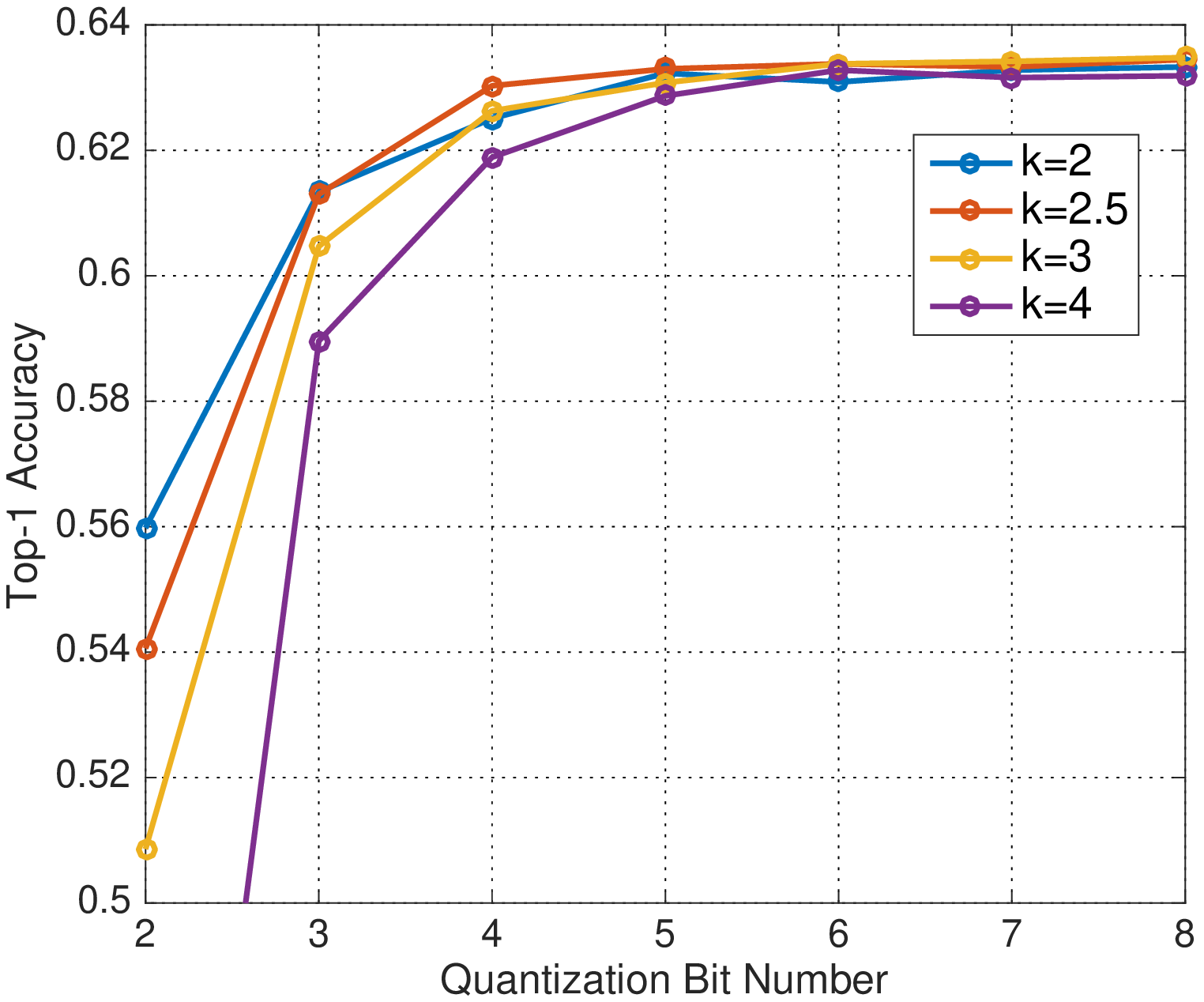}}
    \caption{(a) Quantized-loss of the first convolutional layer's weights. The regular curve is the $QL$ trained without Scale-Clip. (b) Top-1 accuracy of different $k_w$ with different bit.}
  \label{fig:subfig} 
    \vspace{-0.3cm}
\end{figure}

    In Fig.~\ref{fig:qwlr}, the emph{quantized-loss} tends to decrease when bitwidth increases,
    and for same bitwidth, larger $k_w$ tends to have smaller \emph{quantized-loss}, 
    green curves ($k_w=2$) is lower than other curves($k_w>2$)
    This results shows that uniform quantization indeed reduce the \emph{quantized-loss}.
    
    Fig.~\ref{fig:diffk} presents the Top-1 accuracy of the low-bit models after finetuning with 50 epochs.
    The results are consistent with the results shown in Fig.~\ref{fig:diffk}, that $k_w=2$ promotes the low-bit model to achieve the better final performance.
    Thus we can conclude that restricting weight to be uniform-like outperforms those with Gaussian-like or Laplace-like distribution.
	
	\subsection{Group-based Quantization Validation}\label{Exp: group filter}
	  In this part, we conduct experiments to validate the consistent effectiveness of our Group-based Quantization and GDRQ framework.
	  The experiments are also performed on ResNet-$18$(stride is 1 in first block) and CIFAR-$100$.
	  Based on the trained floating-point ResNet-$18$, 
	  we use Group-based Quantization to cluster the convolutional filters into groups by group sizes $gs=[1,4,16,-1]$,
	  where $gs=-1$ is the special case of layer quantization.
	  Then we respectuvely quantize all convolutional layer's weights into $2$-bit and 3-bit with Group-based Quantiztion and fine tune these low-bit models for $50$ epochs.
	  
	\begin{table}
	    \caption{Top-1 Accuracy (\%) of ResNet-$18$ with $n_w=2$}\label{ta:group-filter quantization}
		\setlength{\tabcolsep}{1.5mm}
		\centering
		\begin{tabular}{c|ccccc}
			\hline
			Fine tune      &  float  & 1  & 4   & 16   & -1\\
			\hline
			without finetuning&   73      &  69.3   &  ~50   & ~20   & ~1\\
			\hline
			After 50 epochs & - & 71.3 & 69.5 & 68.1 & 64.9 \\
			\hline
		\end{tabular}
	\end{table}
	  The overall performance of our Group-based Quantization is shown in Table \ref{ta:group-filter quantization}.
	  For $2$-bit weights, the floating-point model just obtains less than $3\%$ accuracy drop by Group-based Quantization  with $gs=1$,
	  while other group sizes obtain much accuracy drop, even the $2$-bit model fails with quantization by layer.
	  The result of 3-bit is consistent with $2$-bit.
	 \begin{figure}
	\centering
	\includegraphics[scale=0.4]{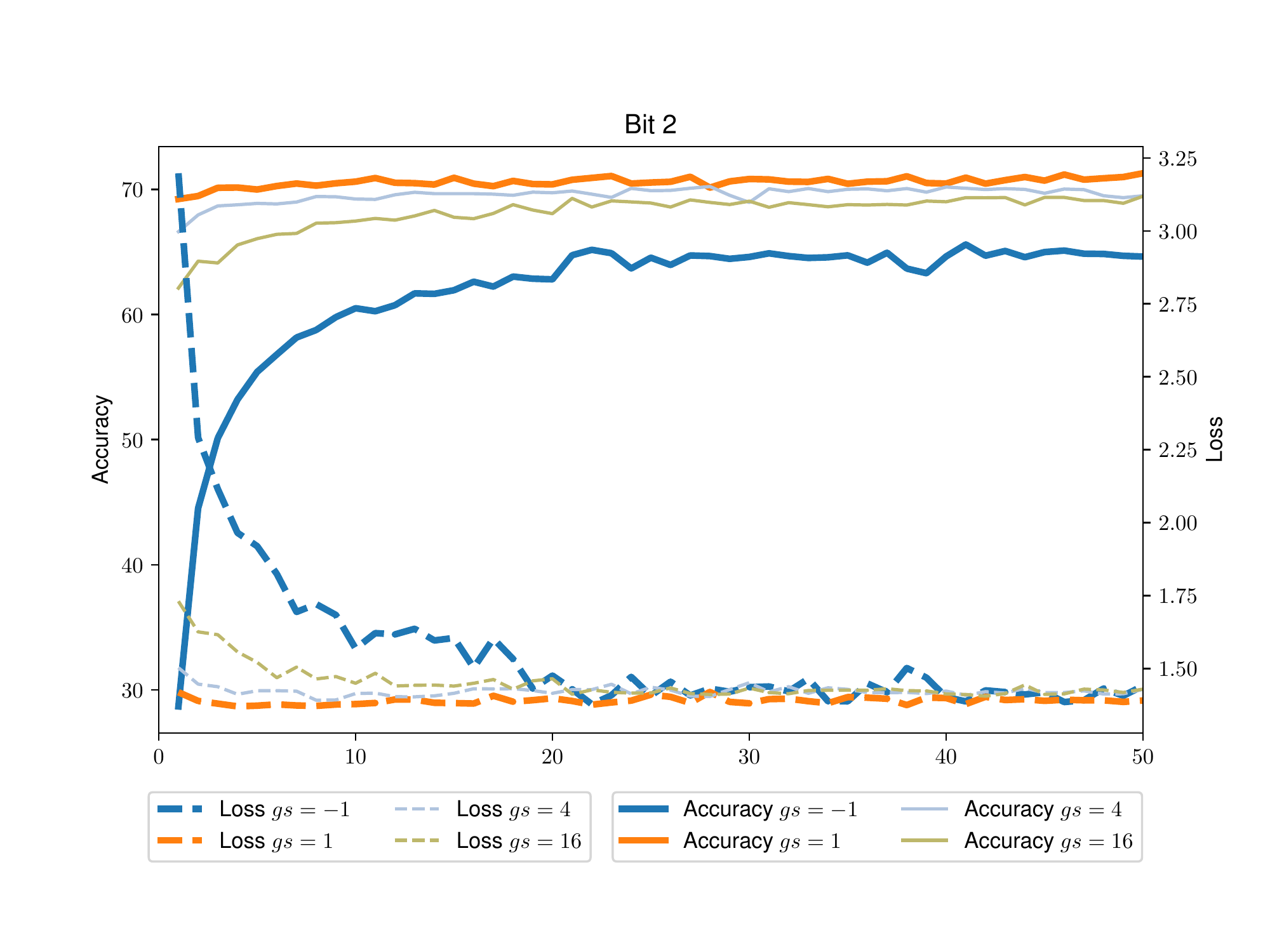}
    \caption{Loss and Accuracy with different group size $gs$ during fine tuning stage. 
    Orange curves denote the loss (dash line) and accuracy of $gs=1$ varies with epochs. 
    Blue curves denote the loss (dash line) and accuracy of $gs=-1$ changing with epochs.}\label{Fig: exp-gr-training}
    \end{figure}
	  After recovering the accuracy drop with finetuning, $2$-bit ResNet-$18$ by Group-based Quantization with $gs=1$ achieves 71.3 with less than 1\% accuracy drop,
	  while $2$-bit quantized ResNet-$18$ by layer, that is $gs=-1$ obtains more than 7\% accuracy drop.
	  The curves in Fig. \ref{Fig: exp-gr-training} also shows that $gs=1$ always achieves better performance than quantization by layer and other group size.
	  Table \ref{ta:group-filter quantization} and Fig. \ref{Fig: exp-gr-training} both show that low-bit models achieve better performance with group-size decreasing.
	  Thus Group-based Quantization can reduce the low-bit model's \emph{quantized-loss} as well as promote the final performance.
	  We believe this is caused by Group-based Quantization increasing the low-bit model's capacity.
	  
	  Further, we impose the Distribution Reshaping on each group filters and quantize ResNet-$18$ into low-bit model.
	  The overall performance of the GDRQ framework is shown in Table \ref{ta:group_size}.
	  
	  \begin{table}
		\caption{Top-1 Accuracy (\%) of ResNet-$18$ on CIFAR-$100$ with GDRQ framework}\label{ta:group_size}
		\centering
		\begin{tabular}{c|cccc}
			\hline
			group size   &  1  & 4   & 16   & -1 \\
			\hline
			float & 73 & 73 &  73 & 73 \\
			$2$-bit&  0.1   &  0.1   & 0.1   & 0.1  \\
		    Binary & -1.6  & -1.4  & -1.4 & -1.7  \\
			\hline
		\end{tabular}
	\end{table}
	The floating-point models trained with different group size achieve similar performance.
	This result shows that applying the Distribution Reshaping on group filters doesn't effect the model's performance.
	and even all $2$-bit ResNet-$18$ have little accuracy drop.
	To compare the the performance with different group size, we further binarize the weight.
	Group-based quantization also achieve better performance in binarized model.
	However, small group size is not always better, since when group size is equal to $1$, the binarized model doesn't outperform other models.
	We think this is because that when imposing the Distribution Reshaping on too small number of weights, the distribution will be unstable.
	Thus, we suggest we should choose the proper group size between increasing low-bit model's capacity 
	and keeping the stable statistical measures for Distribution Reshaping.
	
	\subsection{VGG-$16$ \& ResNet-$50$ on ImageNet}\label{Exp: Networks}
	We quantize two typical CNN models using our Group-based Distribution Reshaping Quantization framework: VGG-$16$ and ResNet-$50$,
	which represents two different CNN architectures respectively.
	Both models are fine-tuned on the ImageNet dataset (ILSVRC-12). 
	Top-1 and Top-5 classification performance are reported on the 50k validation set.
	\vspace{-0.2cm}
	\paragraph{VGG-$16$ on ImageNet}
	As described previously, we impose Distribution Reshaping method by group to train the floating-point model.
	To shape the distribution of activation layer, we also add the Distribution Reshaping in ReLU layers similar to ~\cite{cai2017deep}.
    We use SGD with mini-batch size of 512, and other parameters are kept as the original VGG paper.
    After 50 epochs training, we quantize all convolutional layers' weights and activations of floating-point model except for the first layer.
    Then we fine tune these low-bit VGG-$16$-BN to obtain the final low-bit model.
    
    We summarized the performance of the proposed framework on Table \ref{ta:vgg_ima}.
    Note that we compare the bitwidth of $n_a$ and $n_w$ with $[2,4]$ bit, $[4,4]$bit, $[2,8]$ bit, $[4,8]$ bit, 
    since these bitwidths are more practical.
	\begin{table}
		\caption{Top-1 Accuracy (\%) of VGG-$16$-BN in different bit width. [2,4] denotes $2$-bit for weights and 4-bit for activations.}\label{ta:vgg_ima}
		\begin{tabular}{c|cccccc}
			\hline
			Model      &  float &[2,2]  & [2,4]   & [4,4]   & [2,8] \\
			\hline
			Ours&    72.6  & 69.8   &  71.7   &  72.5   & 72.3  \\
			\hline
		\end{tabular}
	\end{table}
	In Table \ref{ta:vgg_ima}, compared to floating-point model, the low-bit VGG-$16$-BN with 4-bit weights and 4-bit activations has little accuracy drop.
	This demonstrates that under our GDRQ framework the 4-bit VGG-$16$-BN can fully hold the performance.
	With lower bitwidth such as $2$-bit weights, the VGG-$16$ gets less than 1\% accuracy drop.
	Even $2$-bit weights and $2$-bit activation could almost reaches 70\%.
	\paragraph{ResNet-$50$ on ImageNet}
	The proposed quantization framework is also effective to compress the ResNet-$50$ architecture, which achieves state-of-art classification accuracy on ImageNet.
	During the process of training floating-point ResNet-$50$,
	 Distribution Reshaping is also implemented by groups.
	As VGG-$16$-BN, we quantize the trained floating-point ResNet-$50$ into $[2,2]$-bit, $[2,4]$-bit, $[4,4]$-bit and $[2,8]$-bit.
	We fine tune the quantized ResNet-$50$ with learning rate starting at $0.1$ and being divided by $10$ every $20$ epochs.
	
	The overall performance of our quantization framework on quantized ResNet-$50$ is shown in Table \ref{ta:imagenet}.
    There should note that the Top-1 accuracy of floating-point ResNet is less than $76\%$, however we implement the ResNet-$50$ without adding Distribution Reshaping but get similar performance. The reason may be that we adopt improper training hyperparameter on multi-GPUs.
    Even so, the low-bit ResNet-$50$ outperform other methods such as $SYQ$ and $FGQ$.
    For example, $[4,4]$-bit and $[2,8]$-bit ResNet-$50$ with our quantization framework obtains 0.3\% accuracy drop, while $SYQ$ obtains more than 3\% accuracy drop.
    Compared to floating-point ResNet-$50$, $[2,4]$-bit still has less than 1\% accuracy drop. 
	\begin{table}
	    \centering
		\caption{Top-1 Accuracy (\%) of ResNet-$50$ with three different models in different bit on ImageNet.}\label{ta:imagenet}
		\begin{tabular}{c|cccccc}
			\hline
			Model      &  float  &[2,2] & [2,4]   & [4,4]  & [2,8] \\
			\hline
			SYQ        &  76      & -    &  70.9   &   -     & 72.3   \\
			FGQ        &  -      & -     &  68.4   & -       & 70.8   \\
			DoreFa-Net &    -     & -    &  -      & 71.4    & - \\
			\hline
			\textbf{Ours} &    74.8  & 70.6    &  73.9   &  74.5   & 74.5  \\
			\hline
		\end{tabular}
	\end{table}
	\subsection{Comparison on PASCAL-VOC Detection}\label{Exp: detection}
	In this section, we conduct our GDRQ framework in 
	detection task with Faster-RCNN on PASCAL-VOC.
	Note that we use ResNet-$50$ as backbone with pretrained model on ImageNet for Faster-RCNN.
	\hspace{-0.5cm}
	\paragraph{Results}
	\begin{table}
		\caption{mAP of PASCAL-VOC.`*' denotes that activations are not quantized.}
		\label{tab:map}
		\begin{tabular*}{8.5cm}{c|ccccc}
			\hline
			Model  & float & [5,8] & [4,8] & [4,4] & [2,4]  \\
			\hline
			Park et al.   & 77.61 & 77.1$^*$ & 77$^*$ & 72.9 & 66 \\
			Yin et al. & 77.46 & 76.99$^*$ & 74.4$^*$ & - & - \\
			\hline
			\textbf{Ours} & 79.0 & 79.0 & 78.8 & 78.5 & 78.3\\
			\hline
		\end{tabular*}
	\end{table}
	
		From Table~\ref{tab:map}, we notice that the mAP of low-bit fixed point models have little degradation compared to the floating-point models, even $[2,4]$ bit models,
	even models with $2$-bit weights and 4-bit activations only drops $0.7\%$ compared to the floating-point model.
	We compare our quantized detection results with \cite{park2017weighted} and \cite{yin2016quantization}. 
	Note that the networks in \cite{park2017weighted} and \cite{yin2016quantization} are modified version of Faster-RCNN as R-FCN.
	And although they adopt non-uniform quantization scheme which takes non-uniform discrete values and has more expressive ability,
	our method is much better than their results since our $[2,4]$ bit model has no any decline.
	
	\subsection{Comparison on Cityscape Segmentation} \label{exp:seg}
	
	In this part, we conduct our Scale-Clip method in segmentation tasks with PSPNet on Cityscapes. Note that we also use ResNet-$50$ as backbone for PSPNet.

	\begin{table}
	    \centering
		\caption{mIoU of Cityscapes.}
		\label{tab:miou}
		\begin{tabular}{c|ccccc}
			\hline
			Model  & float  &  [8,8] & [4,8] & [4,4] & [2,4] \\
			\hline
			Ours & 75.6 & 75.66 & 75.29 & 75.62 & 74.7 \\
			\hline
		\end{tabular}
	\end{table}
	\vspace{-0.4cm}
	\paragraph{Results}
	From Table~\ref{tab:miou}, we can also observe that the mIoU of low-bit fixed point models have little degration,
	even models with $2$-bit weights and 4-bit activations only drops 0.9\% co
	mpared to the floating-point model.
	As for segmentation,  up to our knowledge there is no any open quantization result on large datasets reported, especially in low-bit quantization,
	so we don't compare our segmentation results.

	\section{Conclusion}
	In this paper, we develop a group-based distribution reshaping quantizaiton framework by incoporating our Distribution Reshaping method and Group-based Quantization for uniform quantization. 
	We elaborate experiments in CIFAR-$100$, ImageNet, COCO, VOC, and network in ResNet-$18$, ResNet-$50$, VGG demonstrate our method generalize well to various dataset, tasks and backbone network. We also make new record for ImageNet low-bit quantization state-of-the-art. Our uniform quantization can easily support FPGA deployment. 

{\small
\bibliographystyle{ieee}
\bibliography{ref}
}

\end{document}